\documentclass[journal]{IEEEtran}
\usepackage{cite}
\usepackage{amsmath,amsfonts}

\usepackage{algorithm}
\usepackage{algorithmicx}
\usepackage{algpseudocode}

\usepackage{array}
\usepackage{textcomp}
\usepackage{stfloats}
\usepackage{url}
\usepackage{multirow}
\usepackage{verbatim}
\usepackage{graphicx}
\def\BibTeX{{\rm B\kern-.05em{\sc i\kern-.025em b}\kern-.08em
    T\kern-.1667em\lower.7ex\hbox{E}\kern-.125emX}}
\usepackage{balance}
\usepackage{graphicx}  %插入图片的宏包
\usepackage{float}  %设置图片浮动位置的宏包
\usepackage{subfigure}  %插入多图时用子图显示的宏包
\begin{document}
% \floatname{algorithm}{Algorithm} %算法
\title{Invisible Backdoor Attack with Dynamic Triggers \newline
against Person Re-identification}
\author{
    %Authors
    % All authors must be in the same font size and format.
    \centerline{Wenli Sun, Xinyang Jiang, Shuguang Dou, Dongsheng Li, Duoqian Miao}
    \centerline{Cheng Deng, \emph{Senior Member, IEEE},  Cairong Zhao}
    \thanks{ Wenli Sun, Shuguang Dou, Duoqian Miao and Cairong Zhao are with the Department of Computer Science and Technology, Tongji University, Shanghai 201804, China (e-mail: 2233055@tongji.edu.cn; dousg@tongji.edu.cn; dqmiao@tongji.edu.cn; zhaocairong@tongji.edu.cn).}
    \thanks{ Xinyang Jiang and Dongsheng Li are with Microsoft Research Asia, Shanghai 200232, China (e-mail: xinyangjiang@microsoft.com;
    dongshengli@fudan.edu.cn).}
    \thanks{Cheng Deng is with the School of Electronic Engineering, Xidian University, Xi'an 710071, China (e-mail: chdeng.xd@gmail.com).}}

%Example, Single Author, ->> remove \iffalse,\fi and place them surrounding AAAI title to use it
\iffalse
\title{My Publication Title --- Single Author}
\author {
    Author Name
}
\affiliations{
    Affiliation\\
    Affiliation Line 2\\
    name@example.com
}
\fi

\iffalse
%Example, Multiple Authors, ->> remove \iffalse,\fi and place them surrounding AAAI title to use it
\title{My Publication Title --- Multiple Authors}
\author {
    % Authors
    First Author Name,\textsuperscript{\rm 1}
    Second Author Name, \textsuperscript{\rm 2}
    Third Author Name \textsuperscript{\rm 1}
}
\affiliations {
    % Affiliations
    \textsuperscript{\rm 1} Affiliation 1\\
    \textsuperscript{\rm 2} Affiliation 2\\
    firstAuthor@affiliation1.com, secondAuthor@affilation2.com, thirdAuthor@affiliation1.com
}
\fi
\maketitle

\begin{abstract}

In recent years, person Re-identification (ReID) has rapidly progressed with wide real-world applications, but also poses significant risks of adversarial attacks. 
In this paper, we focus on the backdoor attack on deep ReID models. 
Existing backdoor attack methods follow an all-to-one or all-to-all attack scenario, where all the target classes in the test set have already been seen in the training set. However, ReID is a much more complex fine-grained open-set recognition problem, where the identities in the test set are not contained in the training set. 
Thus, previous backdoor attack methods for classification are not applicable for ReID. 
To ameliorate this issue, we propose a novel backdoor attack on deep ReID under a new all-to-unknown scenario, called Dynamic Triggers Invisible Backdoor Attack (DT-IBA). 
Instead of learning fixed triggers for the target classes from the training set, DT-IBA can dynamically generate new triggers for any unknown identities. 
Specifically, an identity hashing network is proposed to first extract target identity information from a reference image, which is then injected into the benign images by image steganography.
We extensively validate the effectiveness and stealthiness of the proposed attack on benchmark datasets, and evaluate the effectiveness of several defense methods against our attack.
\end{abstract}


\section{Introduction}

\IEEEPARstart{R}{ecently}, deep learning has progressed rapidly and has been widely utilized in a variety of image classification and recognition tasks. The success of deep learning models is highly dependent on the scale of datasets \cite{sun2017revisiting}. 
However, datasets for deep model training are time and money intensive to construct, resulting in a large portion of the algorithm developers opting for third-party datasets, which brings huge security risks of backdoor attacks \cite{gu2017badnets, sun2019can, li2020backdoor, wang2020attack, bagdasaryan2020backdoor}.

Backdoor attacks occur when samples poisoned with backdoor triggers have their labels changed to target labels and added to the training set, causing the model to mis-classify the target labels during the inference stage. 
As shown in Fig.~\ref{intro}, current backdoor attack scenarios for classification tasks can be classified into two categories: all-to-one and all-to-all \cite{gu2017badnets, li2020backdoor, doan2021backdoor, wang2022bppattack, feng2022fiba}.  
In the all-to-one scenario, a single target label is pre-defined and the poisoned images from any category  will be classified as this fixed label.
On the other hand, for the all-to-all scenario, the target label can be any of the classes that appear in the training set, and images can be manipulated to be classified into any chosen classes.  
For both scenarios, the target labels usually need to be pre-determined before poisoning the training set and the backdoor triggers are kept fixed during the attack. 

In this paper, we shift our focus from classification to another important vision task: Person Re-identification (ReID) \cite{li2014deepreid, zheng2015scalable, zheng2016person, li2018harmonious}. 
ReID is a task of matching person images from several camera viewpoints. 
It has wide applications in surveillance, tracking, smart retail, etc, but also faces the threat of backdoor attacks.
While classification models based on deep learning have been proven to be vulnerable to backdoor attacks \cite{gu2017badnets, liu2020reflection, li2020invisible, saha2020hidden, manoj2021excess}, the backdoor attack risk on ReID has not been thoroughly studied. Existing backdoor attacks on classification cannot be used directly for recognition tasks due to the following challenges. 
Firstly,  ReID is a more challenging fine-grained recognition task with significantly more classes (i.e. person identities) compared to conventional classification. 
This means that there are only a dozen images per class, and traditional attacks significantly alter the image distribution, which is very detrimental to stealthiness.  
More importantly, conventional backdoor attack methods generate fixed triggers corresponding to the pre-determined target labels in the training set. 
However, ReID is an open-set recognition problem where training and test sets have distinct identities. 
In the inference stage, the target ID does not exist in the training set, so a new trigger needs to be generated dynamically according to the target ID. 

To tackle the challenges in backdoor attacks on ReID models, we propose a novel all-to-unknown attack scenario, where the target class can be any of the classes in the test set, even if it does not appear in the training set. In contrast to the conventional attack scenario, new backdoor triggers can be dynamically produced for any novel target identity outside the training set, as shown in Fig.~\ref{intro}. 
Specifically, to realize this scenario, given a benign image, the goal is to generate a backdoor trigger containing an unknown target identity specified by a reference image. 
As a result, the attacked ReID model will recognize the poisoned image as the identity of the reference image.

To dynamically generate the backdoor trigger, we propose an identity hashing network to first encode the target identity information in the reference image as an embedding in hamming space, and then inject the  embedding into the benign images in the form of pixel perturbation by image steganography. As a result, two images of different identities with the same invisible trigger will be recognized as the same person, and vice versa. In summary, our contributions are three-fold as follows:
\begin{figure}[t]
\centering
\includegraphics[width=0.75\columnwidth]{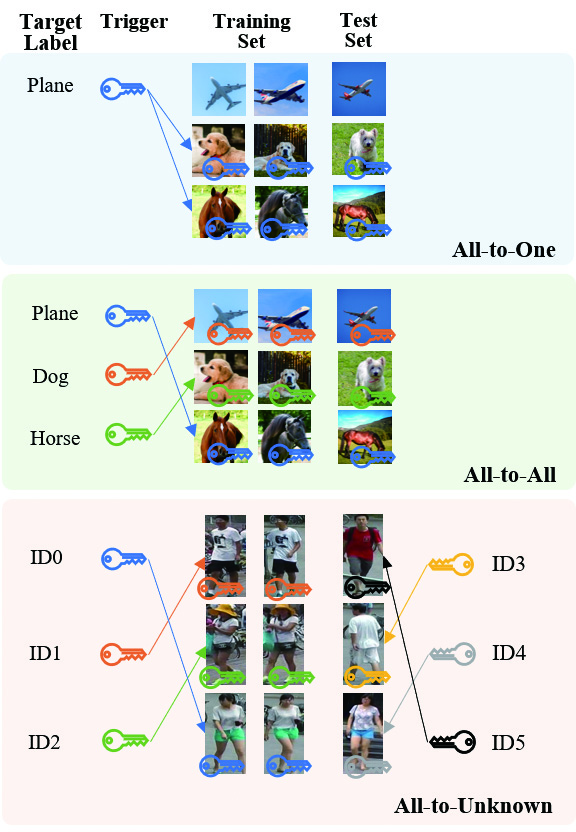} % Reduce the figure size so that it is slightly narrower than the column. Don't use precise values for figure width.This setup will avoid overfull boxes.
\caption{Backdoor attack scenarios for image classification and person ReID. All-to-one and all-to-all are attack scenarios in image classification tasks that require the target labels to be consistent in the training and testing stages. Our proposed all-to-unknown is applicable to the scenario where the target class does not appear in the training set.
}
\label{intro}
\end{figure}

\begin{itemize}
\item We raise a new and rarely studied backdoor attack risk on the ReID task, which is quantified by our proposed new all-to-unknown attack scenario. 
\item  The new all-to-unknown attack scenario and a novel corresponding method are proposed to realize adversary mismatch and target person impersonation by dynamic triggers, which is able to dynamically alter the poisoned image's identity to any target identity outside the training set. 
\item The experiments show the effectiveness of the proposed backdoor attack method and are robust to several existing representative backdoor defense strategies. 
\end{itemize}

\section{Related Work}
In this section, we review current backdoor attacks suggested for image classification tasks, backdoor defense techniques, and ReID models. 
\subsection{Backdoor Attack}
   Backdoor attack is a severe concern to DNNs since they cause the poisoned model to function normally on clean samples but classify samples with triggers as target class \cite{pitropakis2019taxonomy, feng2022fiba, wang2022survey}. 
The majority of existing backdoor attacks are based on the assumption that the target labels are known and fixed, which is only appropriate for classification tasks. Specifically, the adversary needs to design a trigger pattern $t$, and select a target label $y_t$. The adversary adds triggers to benign samples $x\in D$ to generate the poisoned samples $x^p$, then changes the label of $x^p$ to the target label $y_t$ and puts$(x^p, y_t)$ in the training set $D_{train}$. During the training stage, the backdoored model will associate triggers with target labels, resulting in the classification of any sample containing a trigger as $y_t$ during the inference stage. 

Currently,  backdoor attack on classification models is a well-established research field  \cite{DBLP:conf/nips/WangSRVASLP20, zhai2021backdoor, DBLP:conf/ijcai/WangJW0XS21, DBLP:conf/cvpr/ZhaoMZ0CJ20, DBLP:conf/iccv/Xiang0CLK21, li2022few, zhang2022poison}. Recently, few works have shifted their focus to models other than  image classification. Zhai~\textit{et al.}~\cite{zhai2021backdoor} designed a clustering-based attack methodology for speaker verification in which poisoned samples from different clusters include different triggers. All triggers are concatenated with the test samples during the inference stage to achieve the adversary aim. However, because this approach registers only one individual at a time, it is incompatible with recognition tasks requiring a retrieval range greater than one, such as person re-identification. Zhao et al.  proposed a backdoor attack strategy for video recognition models in which the adversary doesn't have to change the label, but is aware of the target class during the training process \cite{DBLP:conf/cvpr/ZhaoMZ0CJ20}.  Few-shot backdoor attack (FSBA) is not a label-targeted attack; instead, it embeds triggers in the feature space, causing the model to lose track of a specific object \cite{li2022few}. To the best of our knowledge, none of the existing backdoor attack methods are applicable to person re-identification. 

\subsection{Backdoor Defense}
Currently, backdoor defense strategies can be classified into the following three categories \cite{li2020backdoor}. 

Sample filtering-based empirical defenses are designed to distinguish between clean and poisoned samples and to use only benign samples for training and testing \cite{ DBLP:conf/iccad/JavaheripiSFJK20, jin2020unified, DBLP:conf/iclr/DuJS20,DBLP:conf/iccv/ZengPMJ21}.
Zeng~\textit{et al.}
revealed the presence of high-frequency artifacts in poisoned images compared to natural images. This approach uses data augmentation to process the training set in order to simulate potential backdoor attack patterns and train a supervised model that distinguishes between poisoned and clean data \cite{DBLP:conf/iccv/ZengPMJ21}. 

Poison suppression-based defenses can be used to eliminate the influence of backdoor triggers during training to prevent backdoor generation  \cite{hong2020effectiveness, li2021anti}.
Hong~\textit{et al.} used differentially private stochastic gradient descent (DP-SGD) to clip and perturb individual gradients during the training stage, ensuring that the generated model contains no hidden backdoors \cite{hong2020effectiveness}.

Model reconstruction-based defenses are aimed at erasing the infected model's hidden backdoor. As a result, even if the trigger remains in the poisoned samples, the prediction stays benign \cite{liu2018fine, wu2021adversarial, zeng2021adversarial, li2021neural}. 
Based on the observation that they are usually dormant in benign samples, pruning backdoor-related neurons was proposed to remove the hidden backdoor \cite{liu2018fine}.

\subsection{Deep Person Re-identification Models}
The goal of ReID is to determine whether a query person has appeared in another place, and was captured by different cameras at a distinct time \cite{gheissari2006person}. 
The pedestrian re-identification models extract robust feature representations of pedestrians through representation learning methods. Then, the similarity score between pedestrians is calculated by the metric learning method and ranked from the highest to lowest, and the target pedestrian is re-identified according to the ranking result \cite{zheng2019pyramidal,zhu2020aware,ye2021deep,zhao2021salience}.

\begin{figure*}[t]
\centering
\includegraphics[width=2.0\columnwidth]{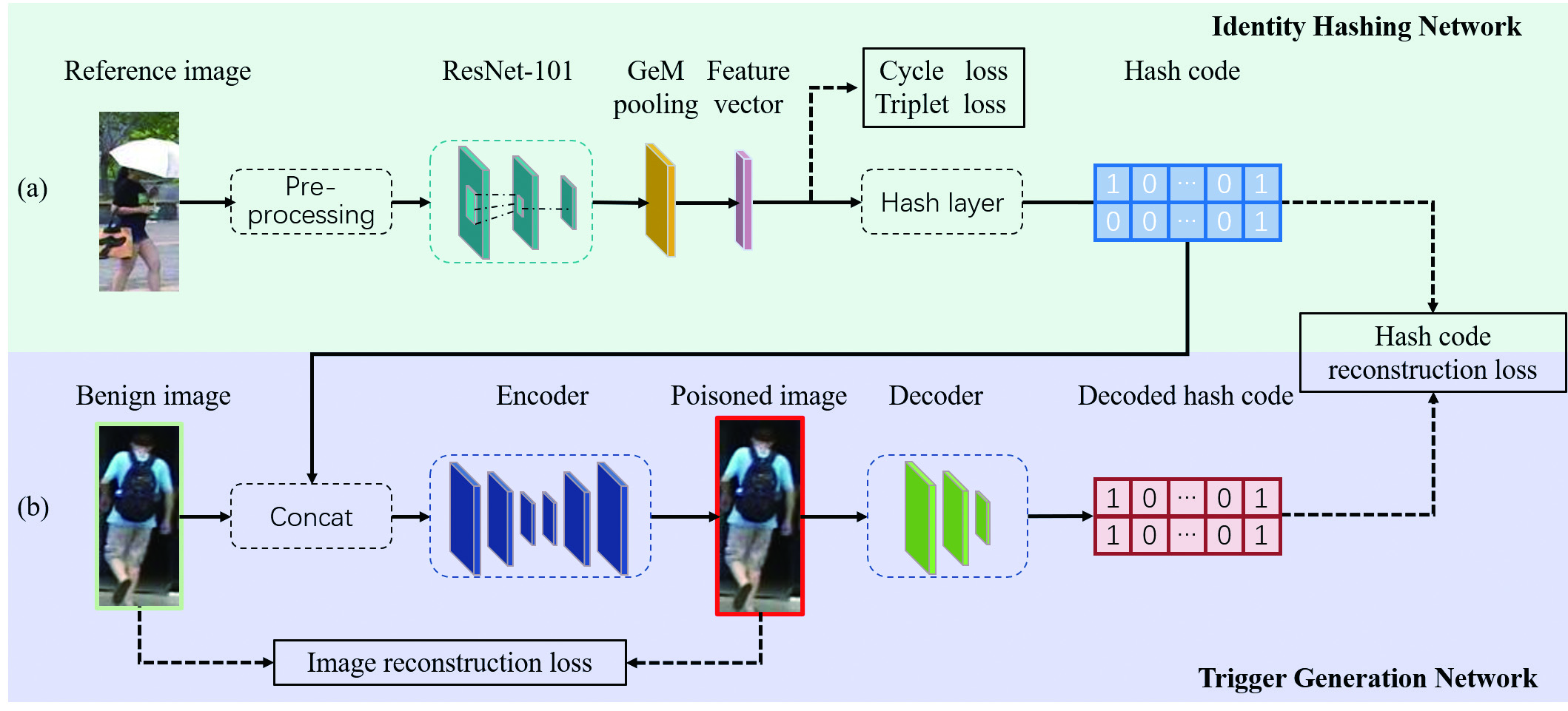} 
\caption{The training pipeline of our poisoned image generator. (a) Identity hashing network. After image pre-processing, the image feature of the reference image is extracted with ResNet-101. And then GeM pooling layer aggregates feature maps generated by ResNet-101. After that, the high-dimensional features are further compressed by the hash layer to obtain a 128-dimensional hash code. 
(b) Trigger generation network. The benign samples are first connected with the hash code of the reference image, and then the deep image steganography network with an encoder-decoder structure is used to generate the trigger. The training stage needs to minimize the perceptual difference between the poisoned and benign images in order to achieve invisibility. And the decoder is trained to reconstruct the hash code.
}
\label{generation}
\end{figure*}
\section{The Proposed Attack}

In this section, we introduce our proposed backdoor attack against person re-identification models under the all-to-unknown scenario. We begin with an introduction to our threat model and an overview of the attack pipeline, followed by an analysis of how to generate and apply the proposed dynamic trigger for the attack.

\subsection{Threat Model}
In this paper, we follow a more realistic setting, where  the adversary gets clean datasets without access to model structure or training loss. 
During the inference stage, the adversary is only permitted to query the trained model with any image. It has no knowledge of the model and cannot manipulate the inference procedure. 
The poisoned dataset is created by dynamically adding imperceptible triggers to some of the training images. 
The aim of the attack is to induce a backdoor behavior in the person re-identification network trained by the user so that it performs normally on a clean test set, but will recognize any person image with a trigger as the target person, regardless of its ground truth.
Therefore this attack is not easily detectable and can escape standard validation tests when users download datasets or models published by the adversary.
The proposed backdoor attack with dynamically generated triggers is able to evade the person search of the deep Re-ID models, causing a serious security risk.

\subsection{Preliminary}

There are two types of attack scenarios for classification tasks  according to the number of target classes~\cite{nguyen2021wanet,wang2022bppattack}. 
 1) All-to-one is a single-target backdoor attack scenario, where the adversary selects a fixed label $c$ as the output label. Eq.~\ref{eq1} shows backdoor generation function $T$ can make $x_i$ mis-classified to the target label $c$.
 \begin{equation}\label{eq1}
 f^{'}(T(x_i))=c  ~~~~\forall (x_i , y_i) \in \mathcal{D}
\end{equation}
 2) All-to-all is a multi-target attack scenario where the target label $\hat{c}$ is the next label of the true label. It is formulated as follows, in which $C$ means the number of classes.
 \begin{equation}\label{eq2}
 f^{'}(T(x_i))=\hat{c} ,~~\hat{c}=y_{(i+1)mod|C|}~~~~ \forall (x_i , y_i) \in \mathcal{D} 
\end{equation}

Since the target IDs of the person re-identification training and test sets are different (\emph{i.e.} $c_{train} \cap c_{test} = \emptyset$), these two conventional attack scenarios are not suitable for ReID. As a result, we propose a new scenario specifically for ReID, called all-to-unknown.

Let $D_{train}=\{(x_{i},y_i)\}_{i=1}^{N_{tr}}$ represents the benign training set containing $N_{tr}$ images, where $y_i \in \{0, 1, ..., M_{tr} - 1\}$ is the $M_{tr}$ person IDs, and $D_{test}=\{(x_{i},y_i)\}_{i=N_{tr}}^{N_{tr}+N_{te}}$ indicates the benign test set containing $M_{te}$ person IDs denoted as $y_i \in \{M_{tr}, M_{tr} + 1, ..., M_{tr} + M_{te} - 1\}$.  %$x_{i} \in \{0, …, 255\}^{C\timesW\timesH} $ .  

The test set $D_{test}$ is divided into two parts: query set and gallery set. For each probe image in the query set, the ReID model $f$ will find $K$ images in the gallery with the highest similarity to it in the inference stage. 

Formally, in the all-to-unknown scenario, given a query image $(x_{i}, y_{i})$, and  a set of gallery images $(x_{t}, y_{t})$, 
a backdoor generator $T(x_i, \hat{y})$ is built to obtain a poisoned image with any target ID $\hat{y}$. 
Given a query image, the backdoor attack aims at ranking gallery images that originally belongs to the same identity out of the top-K list, while ranking  the image pair with different identities but poisoned with the same trigger  within the top-K list: 

    \begin{small}
		\begin{eqnarray}\label{eq3}
		\left\{
		\begin{array}{ll}
		Rank\left( {f^{'}}\left( T(x_{i}, \hat{y}), x_{t} \right) \right) > K,~~ \hat{y} \neq y_i  & \textrm{if } y_i = y_t %~~~~~j,k\in [n] ,~~i \in [N]  %T_{\theta}(H(x_t),x_{i_j})
		\\
        Rank\left( {f^{'}}\left( T(x_{i}, \hat{y}),T(x_{t}, \hat{y}) \right) \right) \leq K \quad &\textrm{if } y_i \neq y_t % ~~~~ j,k\in [n], ~~i,t \in [N]
		\end{array} \right.
	    \end{eqnarray}
	\end{small}
During the training process, the target ID in $T(x_i, \hat{y})$ is selected from $M_{tr}$ IDs in the training set, i.e. $\hat{y} \in \{M_{tr}, M_{tr} + 1, ..., M_{tr} + M_{te} - 1\}$. 
However, in the testing stage, target IDs $\hat{y} \in \{M_{tr}, M_{tr} + 1, ..., M_{tr} + M_{te} - 1\}$, which are all unknown identities during training. 

Consequently, a backdoor attack on the ReID needs to generate new triggers that vary based on the unseen target ID.

\subsection{Backdoor Trigger Generation}
As illustrated in Fig.~\ref{generation}, the invisible triggers generation process consists of two consecutive networks, namely the identity hashing network and the trigger generation network.
Since the target IDs in the inference stage of the recognition task differ from the training stage, the triggers need to be dynamically generated based on the unseen IDs. 
This is achieved by specifying the unseen target ID $\hat{y}$ with a reference image belonging to this ID, denoted as $\hat{x}$.   %as the trigger components.
As a result, we propose an identity hashing network $H$ to extract identity information of $\hat{y}$ from the reference image $\hat{x}$ to obtain a feature representation for this unseen identity. 
Then, we adopt a pre-trained encoder-decoder network $T$ to generate poisoned images dynamically based on the identity feature, following the DNN-based image steganography, StegaStamp \cite{tancik2020stegastamp,li2020invisible}. 
We choose StegaStamp because of its information capacity, stealthiness, and robustness. It can produce a high-quality image with a capacity better in general than other cover-selection and cover-synthesis-based techniques. 
The overall process of generating poisoned images can be formalized as:
\begin{equation}\label{eq4}
  x_{i}^p = T\left(x_i, H(\hat{x};\theta_2) ;\theta_1 \right)
\end{equation}

In this section, we elaborate on the implementation of the aforementioned two networks. 
Specifically, the identity hashing network consists of four modules: image pre-processing, backbone, GeM pooling layer, and hash layer. 
A reference image is first fed into a ResNet-101 backbone to obtain a high-level feature map. 
Then we use GeM pooling  to aggregate the feature maps $\boldmath{X} \in \mathbb{R}^{W\times H\times C}$ into a global feature $g =[g_1, g_2,…,g_c] \in \mathbb{R}^{1\times 1\times C} $, where $W$, $H$, and $C$ denote the width, height, and channel of the feature maps, respectively:
\begin{equation}\label{eq5}
 g_c= {\left( \frac{1}{|\boldmath{X}_c|} \sum\limits_{x \in \boldmath{X}_c} x^{\alpha} \right )}^{\frac{1}{\alpha}}
\end{equation}
where $\alpha$ is a control coefficient. 
At this point, the identity feature of a reference picture has been extracted. However, because the quality of the generated image degrades as steganographic information increases, we employ a hash layer to further compress the feature vector to a much more compact 128-dimensional binary form.

Note that the hash function can be chosen freely based on the length of the generated hash code; in our case, we use MD5 \cite{rivest1992md5}.
We use circle loss and triplet loss to train the identity hashing network.

For trigger generation, we propose to adopt a deep image steganography method \cite{tancik2020stegastamp,li2020invisible} that directly embeds the identity feature in the image. 
Specifically, given an image to be poisoned and the binary hash code of the unseen target identity, we first feed the binary code into a linear layer to generate a $32\times 32\times C$ tensor, which is further upsampled to $H\times W\times C$ and concatenated with the image, where $C$, $W$ and $H$ indicate the channel, width, and height of the input image respectively. 
Then the  $H\times W\times (C\times 2)$ inputs are fed into a U-Net \cite{ronneberger2015u} style encoder to generate the poisoned image,  in which the trigger is an imperceptible pixel-level perturbation.

Image reconstruction loss is applied to ensure that the encoder can embed hash code in poisoned images while minimizing the perceptual difference between poisoned and benign images. As shown in Eq.\ref{eq6}, it includes three losses, which are L2 residual regularization loss $L_R$, LPIPS perceptual loss $L_P$, and critical loss $L_C$ calculated between the encoded image and the original image. Through experimental verification, it is found that the attack performance, as well as stealthiness, are best when the hyperparameters $\lambda_R$, $\lambda_P$, and $\lambda_C$ are set to 1.5, 2, and 1, respectively.

\begin{small}
		\begin{eqnarray}\label{eq6}
		L_{image} = \lambda_RL_R + \lambda_PL_P +\lambda_CL_C
	    \end{eqnarray}
\end{small}

This method can achieve as much invisibility of the trigger as possible while maintaining attack performance and increasing the threat of backdoor attacks. 
To ensure that triggers are added to the images effectively, an extra decoder is added to predict the identity hash code from the poisoned image, and another reconstruction loss is added between the predicted code and the original code generated by the identity hashing network.

\begin{figure*}[t]
\begin{center}
\subfigure[Clean mode]{
\includegraphics[width=0.22\textwidth,height=2.5in]{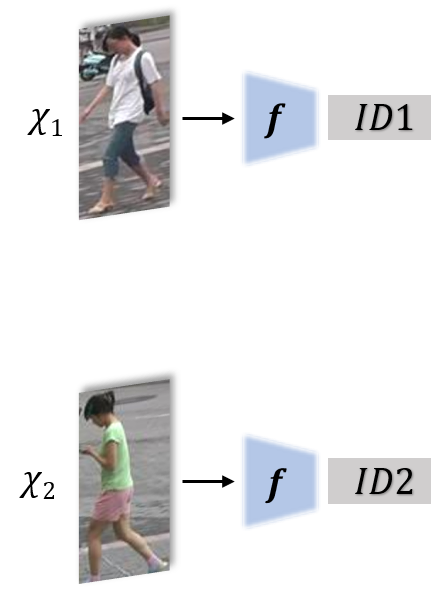}
}
\quad\quad\quad\quad\quad
\subfigure[Target-aware attack mode]{
\includegraphics[width=0.5\textwidth,height=2.5in]{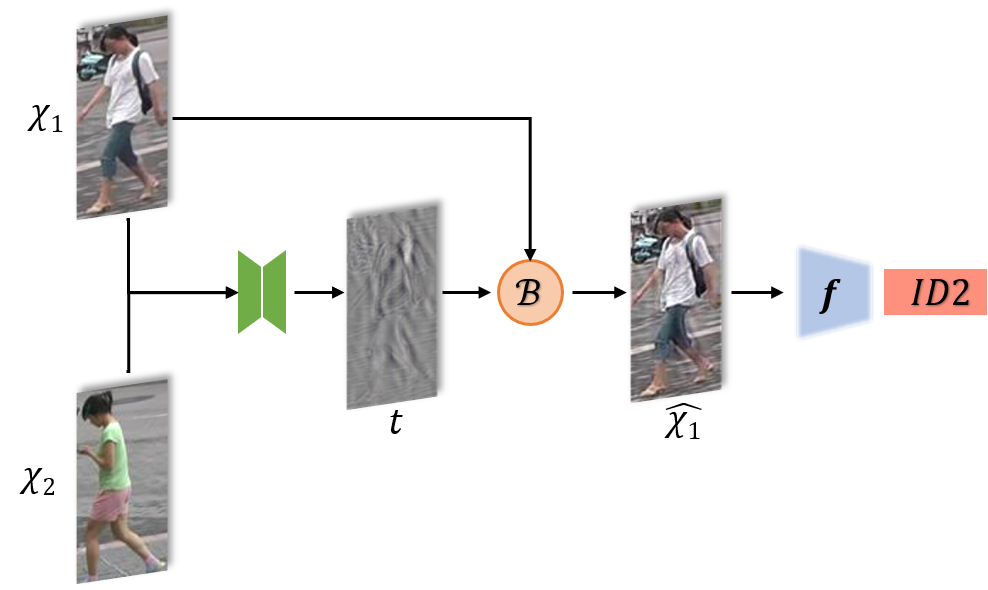}
}
\caption{Two running modes of the proposed target-aware backdoor system. $X_1$ and $X_2$ are benign images of different IDs. In clean mode, the backdoored model can correctly predict the identity.
But in target-aware attack mode, $X_1$ is injected with a trigger related to the target person $X_2$, and then the poisoned image $\hat{X_1}$ is generated.  The trigger activates the backdoor in the model, so $\hat{X_1}$ is predicted as the target ID.
}
\label{mode}
\end{center}

\end{figure*}

\subsection{Backdoor Implantation}

In this section, we introduce how to implant the backdoor trigger into the training set. 
The person ReID dataset is more sparse than  image classification, where the number of images for each ID is extremely low.
The training process follows the all-to-all scenario, in which we use all odd IDs as the target classes of images with even IDs, extract the hash code of each target ID, and inject it into the corresponding training samples by image steganography to generate the poisoned samples.
Finally, the poisoned samples and clean samples are provided to the user for training.
Because the target IDs in the inference phase differ from that in the training stage, the hash code of unseen IDs in the test set is regenerated based on the reference images and injected into the test set. 
When the attacked model gets an image with a trigger as input, it is manipulated to retrieve the images of the same target ID with the trigger. 
The adversary tracked by the ReID model can evade the tracking of the model by injecting a trigger of another one's ID. 
We note that IBA \cite{nguyen2020input} is an all-to-one backdoor attack method that generates backdoor triggers based on the input, and the images are fed into an encoder that can generate specific patterns. It differs from the classical backdoor attack by grouping the backdoor behavior into three modes: clean mode in which the network can correctly identify clean samples; attack mode in which the backdoor is activated when poisoned data is input; cross-trigger mode in which different inputs do not generate the same triggers, and the triggers of different images cannot be applied to another image. 
As shown in Fig.~\ref{mode}, the difference between ours and IBA is that we generate target-aware triggers, and the poisoned images with different triggers can be recognized as the same target person.
The triggers are generated dynamically based on the benign images and target images, and are not determined by the benign images only.

\section{Experiments}
In this section, we evaluate the effectiveness and stealthiness of our proposed backdoor attack on the person ReID models and its resistance to several backdoor defense methods. We conduct ablation studies to demonstrate the importance of our design and to justify the choice of hyperparameters.
\subsection{Experimental Settings}
\paragraph{Datasets and DNN Models}We choose two benchmark datasets for person re-identification: Market-1501 \cite{zheng2015scalable} and DukeMTMC-reID  \cite{zheng2017unlabeled}  to evaluate our attacks.  Market-1501 contains 36036 images of 1501 person IDs. 751 IDs were labeled for the training set and 750 IDs were labeled for the test set. In DukeMTMC-reID, a total of 36,411 images of 1812 pedestrians were collected, with 702 IDs in each of the training and test sets. 
To test the attack performance, we select four victim ReID methods (FastReID \cite{wang2018learning}, BoT \cite{luo2019strong}, AGW \cite{ye2021deep}, MGN \cite{wang2018learning}),  each with ResNet-50\cite{he2016deep} and ResNet-101\cite{he2016deep} backbones for training. 
In addition, the adversary cannot acquire any information about the target ReID model. In fact, the feature extractor of the identity hashing network is constructed and trained independently, which is not the same as that of ReID, and the two are unrelated.  The target model is set to different ReID methods, while the identity hashing network is kept the same.
And the poisoned image generator trained on the same dataset remains consistent with the parameters that do not change with the target model.

\paragraph{Baseline}
Given the complex nature of target labels in all-to-unknown scenarios, there is currently no backdoor attack that can be utilized directly against Person ReID. 
Existing attack methods cannot handle the case where the target label is not contained in the training set. Therefore, we only compare with existing methods in terms of the performance of non-targeted attacks where impersonating a specific target identity is not required (more details in evaluation metrics). 
We select Bednets \cite{gu2017badnets}, Blended \cite{chen2017targeted}, ReFool \cite{liu2020reflection}, SIG \cite{barni2019new}, and WaNet \cite{nguyen2021wanet} as our baseline. 
Specifically, Badnets is a patch-based backdoor attack that generates poisoned images by adding pixel patterns to benign images. 
The trigger of Blended is a picture unrelated to the poisoned image, and the poisoned images are generated by overlaying the trigger picture and the original images in a certain ratio. 
ReFool plants reflections that are also images outside the training set as backdoor triggers into a victim model.
SIG uses a ramp signal as a trigger for the poisoned image, which is perceptually invisible.
And WaNet is an invisible backdoor attack that injects backdoors by image distortion. In order to improve the ASR, a noise mode is proposed and trained simultaneously with the clean mode and attack mode.
All of these methods generate poisoned images and put them into the training set, resulting in the backdoored model to mis-classify images with triggers in the test stage. 

% \begin{table}[t]\small
% \renewcommand{\arraystretch}{1.2}
% \centering
% \caption{The attack effectiveness and stealthiness of  several classical backdoor attack models and ours to attack FastReID on  Market-1501 dataset. The ‘BA’ and ‘R-10’ indicate the benign accuracy and the rank-10 accuracy on poisoned data, respectively. And the remaining three metrics SSIM, PSNR, and LPIPS are used to measure the similarity to the original image. 
% }
% %\resizebox{.95\columnwidth}{!}{
% \begin{tabular}{c|ccccc}
% \hline
%     Attack&BA$\uparrow$&R-10$\downarrow$&SSIM$\uparrow$&PSNR$\uparrow$&LPIPS$\downarrow$\\\hline
%     BadNets &\textbf{98.40}&\underline{2.02}&\textbf{0.9736}&20.91&0.0359\\\hline
%     Blended &\textbf{98.40}&2.43&0.6995&19.49&0.0597 \\\hline
%     SIG &97.80&60.57&0.6135&\underline{21.68}&\textbf{0.0153} \\\hline
%     ReFool&\underline{98.16}&17.96&0.8076&20.21&0.0903\\\hline
%     ours &98.10&\textbf{1.07}&\underline{0.8315}&\textbf{26.46}&\underline{0.0203}\\\hline
   
% \end{tabular}

% \label{table2}
% \end{table}

\begin{table*}[t]\small
\renewcommand{\arraystretch}{1.1}
\centering
\caption{The performance (\%) of different Person ReID models under no and our attacks on Market-1501 and DukeMTMC-reID.}
\begin{tabular}{c|cc|cccc|cccc}
    \hline
   \multirow{2}{*}{Methods}&\multirow{2}{*}{Backbone} &\multirow{2}{*}{Model} & \multicolumn{4}{c|}{Market-1501} & \multicolumn{4}{c}{DukeMTMC-reID} \\
    \cline{4-11}
    &&& BA$\uparrow$&ASR$\uparrow$&R-10$\downarrow$&mAP$\downarrow$
    &BA$\uparrow$&ASR$\uparrow$&R-10$\downarrow$&mAP$\downarrow$\\ \hline
    \multirow{4}{*}{FastReID}&\multirow{2}{*}{ResNet50}&Clean&98.99&-&98.43&72.92&96.77&-&89.00&54.15 \\
    &&Backdoored&98.10&97.98&1.07&0.23&96.10&92.77&0.27&0.26   \\
     &\multirow{2}{*}{ResNet101}&Clean&99.14&-&98.87&84.65&96.63&-&89.95&57.92 \\
    &&Backdoored&97.18&91.86&0.50&0.48&91.56&80.43&4.67&1.16\\\hline
     \multirow{4}{*}{BoT}&\multirow{2}{*}{ResNet50}&Clean&98.78&-&98.13&79.80& 95.92&-&87.34&52.47\\
     &&Backdoored&98.72&99.91&0.15&0.44&95.47&93.36&0.58&0.37 \\
     &\multirow{2}{*}{ResNet101}&Clean&98.81&-&98.10&81.94&96.41&-&88.73&54.55 \\
    &&Backdoored&98.60&99.91&0.21&0.42&95.96&93.04&0.63&0.34 \\\hline
    \multirow{4}{*}{AGW}&\multirow{2}{*}{ResNet50}&Clean&98.63&-&98.01&79.95& 95.47&-&86.27&52.90\\
     &&Backdoored& 98.99&99.91 &0.15&0.41&95.92&93.36&0.18&0.26 \\
     &\multirow{2}{*}{ResNet101}&Clean&98.72&-&97.83&80.57&95.74&-&88.11&55.10 \\
     &&Backdoored&98.87&99.91&0.09&0.44&96.01&93.36&0.27&0.28 \\\hline
    \multirow{2}{*}{MGN}&\multirow{2}{*}{ResNet50}&Clean&98.72&-&98.01&65.89&93.77&-&83.12&43.15\\
     &&Backdoored&97.71&99.91&0.15&0.44&93.09&86.31&0.54&0.26 \\ \hline
    %  \multirow{2}{*}{ViT}&Clean Model&&&&&& \\
    % &Backdoored Model&&&&&& \\\hline
\end{tabular}

\label{table1}
\end{table*}

\begin{algorithm}[h]
    \caption{Calculate the ASR of a given poisoned test set} 
    \begin{algorithmic}[1] 
        \Require poisoned test set {$D_t$} with gallery {$G$} and query image {$Q$}, Ground-truth ID annotation $Y_{gt}$, Target ID annotation $Y_{tar}$ % , Ground-truth annotation of tr
        \Ensure $ASR$
        \State $queryCnt$ = 1,  $attackSuccQuery$ = 1, 
          $targetedFlag$ = $True$, $attackFailFlag$ = $False$
         %\hfill $\rhd$ {$attackFailFlag$ : indicates whether it is a target attack or non-target attack.} 
        \For{$query_i$ in $Q$} 
            \State $Ir_i = ReID (query_i)$  \hfill$\rhd${$Ir_i$ : top-10 retrieval images from $G$.} 
            \State $queryCnt++$
            \State $attackFailFlag = False$
             \For{$i$ in $Ir_i$}  \hfill $\rhd${$i$ : iterate through all retrieved images.}   
                \If {$targetedFlag == False$}
                    \If {$Y_{gt}[query_i] == Y_{gt}[i]$}
                        \State $attackFailFlag = True$
                        \State \textbf{break}
                    \EndIf        
                \ElsIf {$Y_{tar}[query_i] == Y_{gt}[i]$}
                    \State $attackSuccQuery$++
                    \State \textbf{break}
                \EndIf
                \EndFor
            
            \If {$targetedFlag == True$} 
                \State Continue
            \ElsIf{$attackFailFlag == Flase$}
                \State $attackSuccQuery$++   
            \EndIf    
        \EndFor
        \State $ASR = attackSuccQuery / queryCnt$\\
        \Return{$ASR$} 

    \end{algorithmic}
\end{algorithm}

\paragraph{Evaluation Metrics}
Assuming that the ReID model retrieves 10 pedestrian images  most likely to be of the same identity as the query image, the success of attacking a ReID model has two criteria. 
The first is a non-targeted attack criterion (i.e.evading attack), which measures attack method's ability to manipulate the target model to rank the positive images outside the top-10 list. 
Specifically, we use the retrieval performance  rank-10 (R-10) and mean Average Precision (mAP) on the poisoned images as the metrics, where lower rank-10 and mAP indicate better non-targeted attack performance. 
The second criterion is targeted attack (i.e.impersonation attack), which adds a condition to the first one. For  a targeted attack, the adversary assigns a specific target person to retrieve, and an attack is only counted as successful when the target person appears in the top-10 rank list \cite{fang2022backdoor,wang2019advpattern}. Following existing works  \cite{nguyen2021wanet,wang2022bppattack,zhao2022defeat,fang2022backdoor}, we use attack success rate (ASR) (i.e. proportion of successful attacks) as the performance metrics of targeted attack. 
The ASR calculation given a test set is detailed in Algorithm 1.
Here, if $targetedFlag$ is $True$, the ASR of the targeted attack is evaluated; otherwise, the ASR of the non-targeted attack is evaluated. In general, the filename of each image in the dataset contains the ground-truth ID.  
For each query, the ReID model retrieves ten images with the same identity in the gallery. If one of these ten images belongs to the ground-truth ID, $attackFailFlag$ is set to true, indicating that the non-targeted attack failed once. 
And if one of the ten images belongs to the target ID, it is counted as a successful targeted attack.
Moreover, we use benign accuracy (BA) to evaluate whether the attack model can perform normally on clean data, which is the percentage of clean probe images that successfully ranks the positive image in the top-10 list.
To evaluate the stealthiness of the backdoor triggers, we select the metrics: structural similarity index (SSIM) \cite{wang2004image}, peak-signal-to-noise-ratio (PSNR) \cite{huynh2008scope}, and learned perceptual image patch similarity (LPIPS)  \cite{zhang2018unreasonable} to measure the differences between clean and poisoned images.

\begin{figure*}[t]
\centering
\includegraphics[scale=0.6]{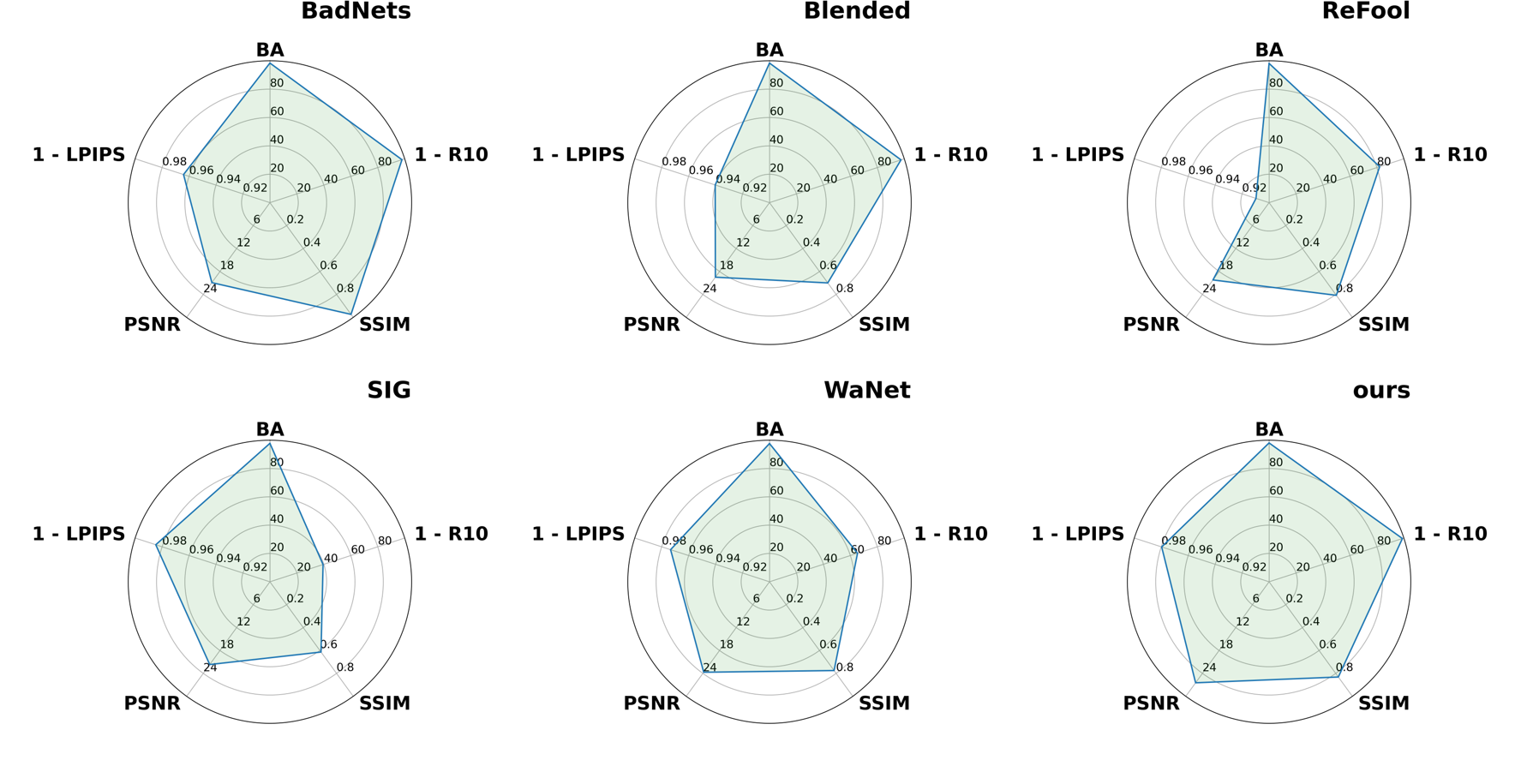} 
\caption{The attack effectiveness and stealthiness of  several classical backdoor attack models and ours to attack FastReID on  Market-1501 dataset. The ‘BA’ and ‘R-10’ indicate the benign accuracy and the rank-10 accuracy on poisoned data, respectively. And the remaining three metrics SSIM, PSNR, and LPIPS are used to measure the similarity to the original image.  Subtracting LPIPS and R-10 from 1 brings them in line with the rest of the metrics, with larger indicating better performance.
}
\label{radar}
\end{figure*}

\subsection{Attack Results}

\paragraph{Effectiveness}
Table~\ref{table1} shows the attack results of different ReID models on Market-1501 and DukeMTMC-reID datasets at a poisoning rate of $37.5\pm1.5\%$.  
In fact, the ASR of non-targeted attacks is obtained by subtracting BA from 100\%, so we do not compare this metric separately in this table.
It is shown that our attack method is able to achieve a high ASR over different ReID methods and different backbones on two standard benchmark datasets. Moreover, we observe a significant performance decrease of the backdoored model on poisoned data in terms of both rank-10 and mAP, further validating our method's effectiveness on the non-targeted attack. 
In addition, we also evaluate the ReID performance of models trained on clean data and observe that there is only a very small gap between its performance on clean and poisoned data.

Figure~\ref{radar} shows the performance comparison between our method and existing backdoor attacks against FastReID at a 37.5\% poisoning rate. 
Note that in Fig.~\ref{radar} only non-targeted metric rank-10 is compared because the target individuals in the training stage cannot be specified by the other methods, making the calculation of the targeted ASR impossible. 
To allow all dimensions to positively reflect attack performance, we subtract the two metrics R-10 and LPIPS from 1, respectively.
In Fig.~\ref{radar}, it is shown that our proposed attack gives the model the lowest rank-10 accuracy of 1.07\% on the poisoned data at the cost of a very small accuracy loss ($0.3\%$ compared to BadNets).

\begin{figure}[t]
\centering
\includegraphics[width=0.95\columnwidth]{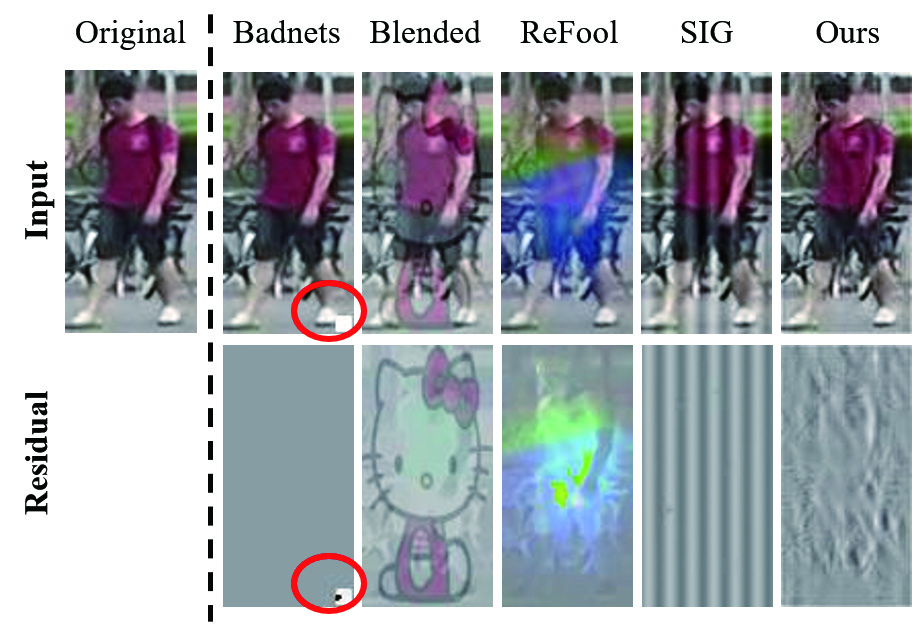} 
\caption{The Display of poisoned images. Top: the original image; poisoned images generated by Bednets, Blended, ReFool, SIG, and ours. Down: the residual images.  
}
\label{demo}
\end{figure}

\paragraph{Stealthiness}
 As shown in Fig.~\ref{radar}, we evaluate the stealthiness of the poisoned image by measuring the difference between the original and poisoned image in terms of SSIM, PSNR, and LPIPS. In terms of ASR metrics, our method is the first to successfully attack in an all-to-unknown scenario, and all other methods are not applicable to the scenario. Thus we only compared stealthiness and ReID performance degradation with them in Fig.~\ref{radar}.
It can be observed that our method is ranked top two in terms of every stealthiness metric and achieves the best trade-off between BA, Rank-10, and stealthiness. Specifically, Fig.~\ref{radar} shows ours achieves the maximum performance drop on the poisoned images with a slight BA drop, which is a significant improvement.

From the visualization in Fig.~\ref{demo}, we can clearly observe the changes made to the poisoned image by the previous methods, while for our method the change is almost not perceivable. 
Note that, Badnets has a high SSIM because the changes to the image are only in the lower right area, while SSIM indicates the average difference of all pixels, but the trigger produced by Badnets is humanly perceivable in the bottom right corner of the image as shown in Fig.~\ref{demo}.

\begin{figure*}[t]
\centering
\includegraphics[width=2.0\columnwidth]{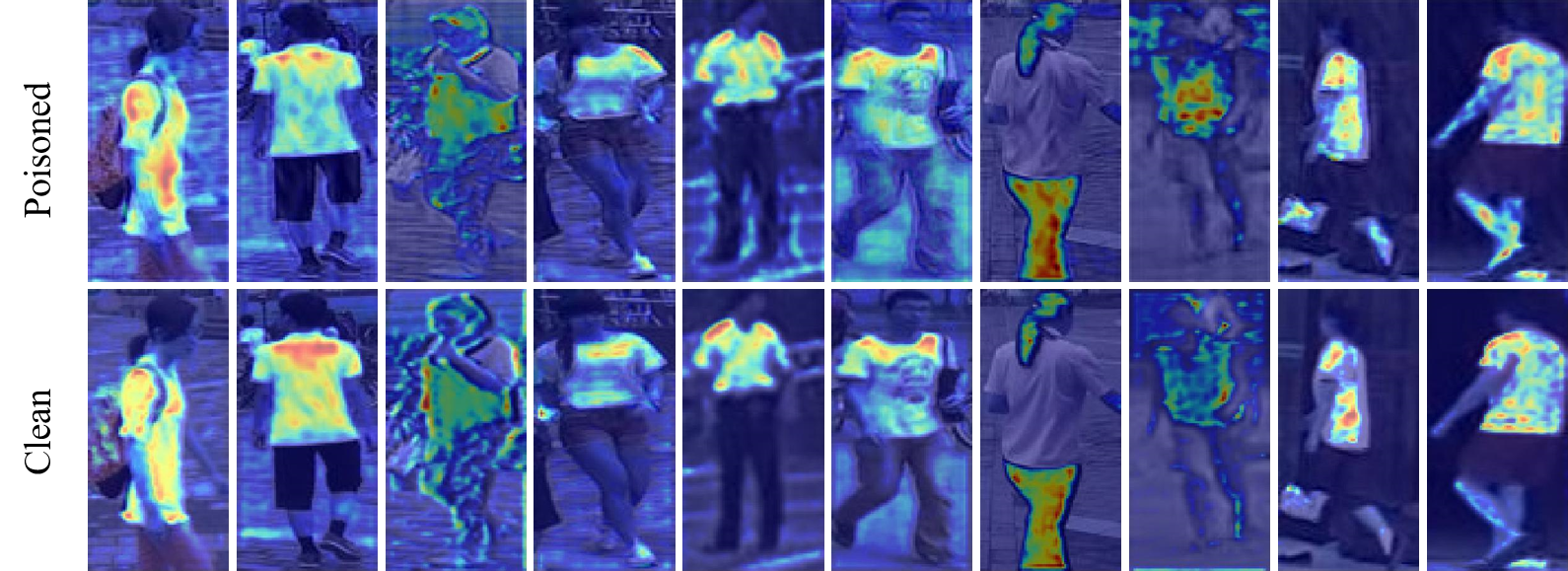}
\caption{
An illustration of Grad-CAM for the samples in the Market-1501 dataset, where the higher luminance regions are the hot spots and contribute the most to the ReID model.
}
\label{grad-cam}
\end{figure*}

\begin{figure}[h]
\centering
\includegraphics[width=0.95\columnwidth]{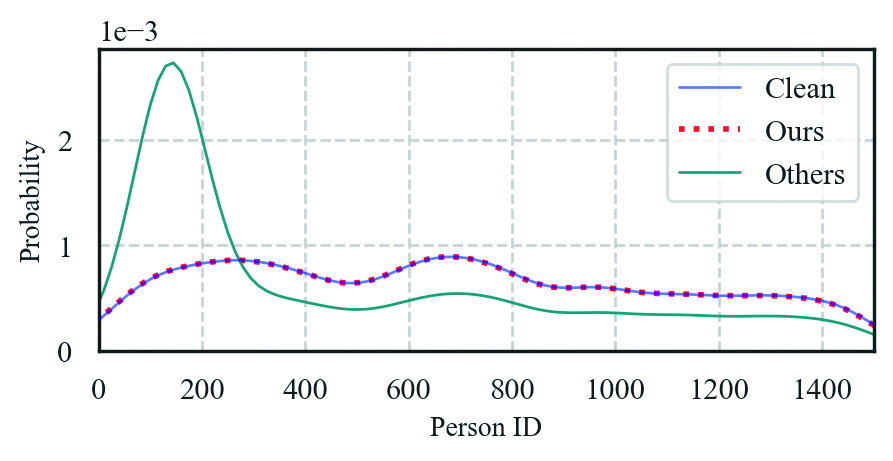}
\caption{Data distribution over different identities. The distribution of the training set data after poisoning Market-1501 with the traditional backdoor attacks and ours ($\gamma =38.8\%$), respectively.
}
\label{distribution}
\end{figure}

 As shown in Fig.~\ref{distribution}, the blue solid line represents the probability distribution of the original clean data, and the green solid line represents the distribution of the traditional attacks on the training set after poisoning. It can be observed that the distribution of the training set is significantly altered due to changing the labels of many poisoned images to the target labels. While the red dashed line represents our attack, the distribution of data after poisoning is almost the same as the clean training set. Specifically, previous backdoor attack methods usually produce a significant and easily detected change to the training data distribution, because they require selecting a target class during the training stage and then changing the label of the poisoned data to the target label, resulting in a significant increase in the number of images in a  particular identity. 
The data distribution on different identity changes produced by previous methods changes drastically compared to the clean data, which can be detected easily and hence loses stealthiness. 
On the other hand, our method only needs to add a few images to each target identity in the whole dataset, so it hardly changes the data distribution.

\subsection{Resistance to Backdoor Defense Methods}
\paragraph{Resistance to RBAT}
In reality, person re-identification models often employ various data augmentation methods that may remove or corrupt backdoor triggers in the post-poisoned images. 

\begin{table}[h]\small
\renewcommand{\arraystretch}{1.2}
\centering
\caption{ The results of resistance to RBAT}
%\resizebox{.95\columnwidth}{!}{
\begin{tabular}{ccc}
\hline
    Datasets&Clean Accuracy&Poisoned Accuracy\\\hline
    Market1501&99.46&1.31\\
    DukeMTMC-reID &99.87&2.74\\\hline
   
\end{tabular}

\label{table3}
\end{table}

To test whether our attack can break through the RABT \cite{DBLP:conf/iccv/ZengPMJ21} defense, we train the defense model on the clean Market-1501 training set and DukeMTMC-reID training set, respectively, and both converge after 20 epochs. Then we embed triggers to  images in test sets, where the target ID of an image with an odd ID is set to an even ID, and vice versa. 
In the test stage, the poisoned images and the original images in a 1:1 ratio are fed into to the pre-trained RABT model, which outputs confidence of the clean and poisoned samples. As shown in Table \ref{table3}, the RABT model can discriminate clean samples with close to 100\% accuracy, while the accuracy of poisoned samples does not exceed 3\% on both datasets. In other words, RABT classifies 98.69\% of the poisoned data as benign samples on Market-1501, and 97.26\% on DukeMTMC-reID. It can be seen that our dynamically generated triggers can bypass the defense of RABT. 

\begin{figure}[t]
\centering
\subfigure[Market-1501] 
{\includegraphics[width=0.85\columnwidth]{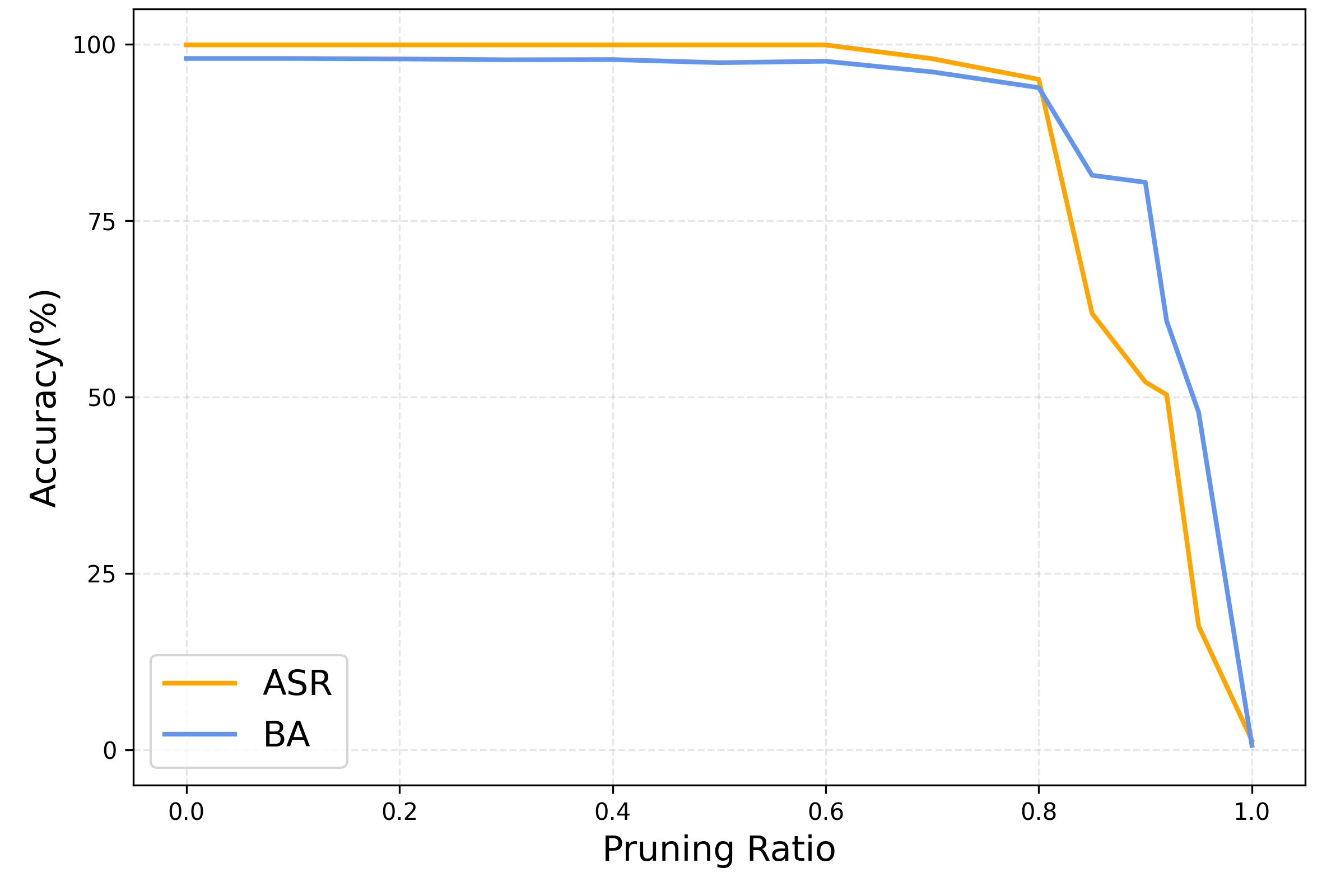}}\\
\subfigure[DukeMTMC-reID]{
 \includegraphics[width=0.85\columnwidth]{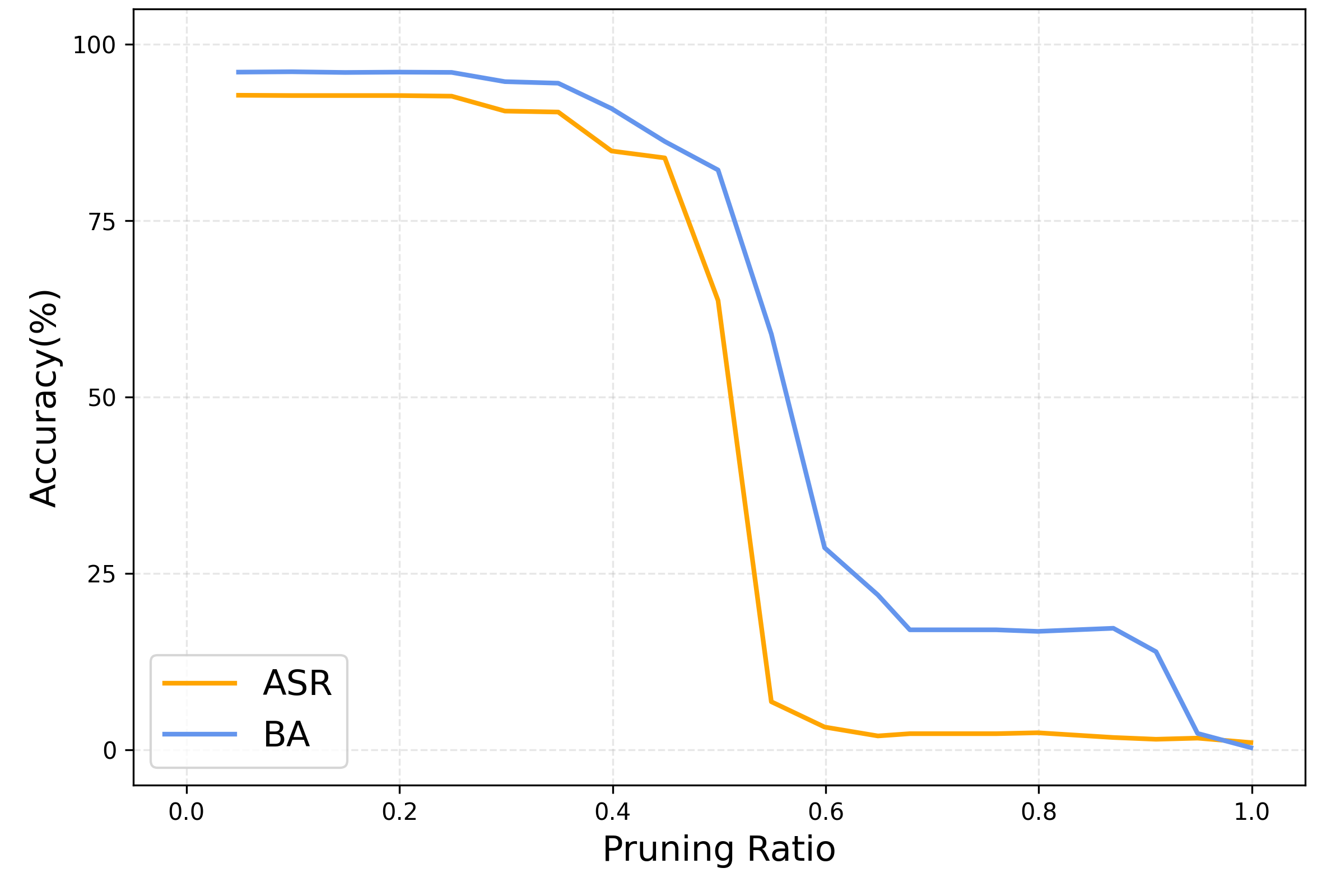}
}
\caption{The results of resistance to Fine-pruning on Market-1501 and DukeMTMC-reID datasets. In both subgraphs, the ASR decreases in parallel with the benign accuracy as the proportion of pruned neurons increases, making it difficult to defend against our attacks by Fine-pruning.
}
\label{fp}
\end{figure}

\paragraph{Resistance to Fine-pruning} The purpose of fine-pruning is to remove backdoor-related neurons that are normal in benign images but are abnormally active in poisoned images \cite{liu2018fine}.
We use the fine-pruning algorithm to the person ReID to prune the neurons activated by the backdoor in the last two layers of the first and the last backbone modules.
We follow the prune-finetune-then-test pipeline and increase the number of the pruned neurons in the designated layers until only 5\% of neurons remained.
It can be found  that fine-pruning slightly reduces the attack success rate, but the model's performance on clean data decreases at a similar rate, as shown in Fig.~\ref{fp}. 

\paragraph{Resistance to Februus}With Grad-CAM \cite{selvaraju2017grad}, Februus \cite{doan2020februus} visualizes the activation hot spots in the penultimate layer of a DNN and replaces the pixels corresponding to the highest-scoring hot spots with image patches recovered by a pre-trained GAN that is used to remove potential backdoor triggers.
We first provide the hot spots map in Fig.~\ref{grad-cam}. The hot spots (shadows with higher luminance) in the poisoned images overlap with the body or clothing of pedestrians, which are almost no different from the clean images. In addition, the triggers generated by our attack are spread over the whole images and cannot be defended by replacing local image patches. We conduct this defense experiment on the Market-1501 dataset and found that our attack can still maintain 93.75\% ASR after trigger pattern removal.

\subsection{Ablation Studies}

\paragraph{Importance of our design}
In this experiment, we take Market-1501 dataset as an example to ablate our design. First, we replace the hashcode generated by the identity hashing network with random noise to create poisoned images. However, if random noise is used as the input of the identity hashing network instead of the target pedestrian features, it does not reduce the distance between the poisoned image and the target person in the feature space as our proposed method does. We conduct the ablation experiment to validate this point, and the ASR when the poisoning rate = 39\% in the test set is only 1.3\%, and the ASR when the poisoning rate = 49\% is only 29.6\%, which is enough to prove that the selection of reference image is necessary.

\begin{figure}[t]
\centering
\subfigure[The ASR and BA w.r.t. different poisoning rates.] 
{\includegraphics[width=0.975\columnwidth]{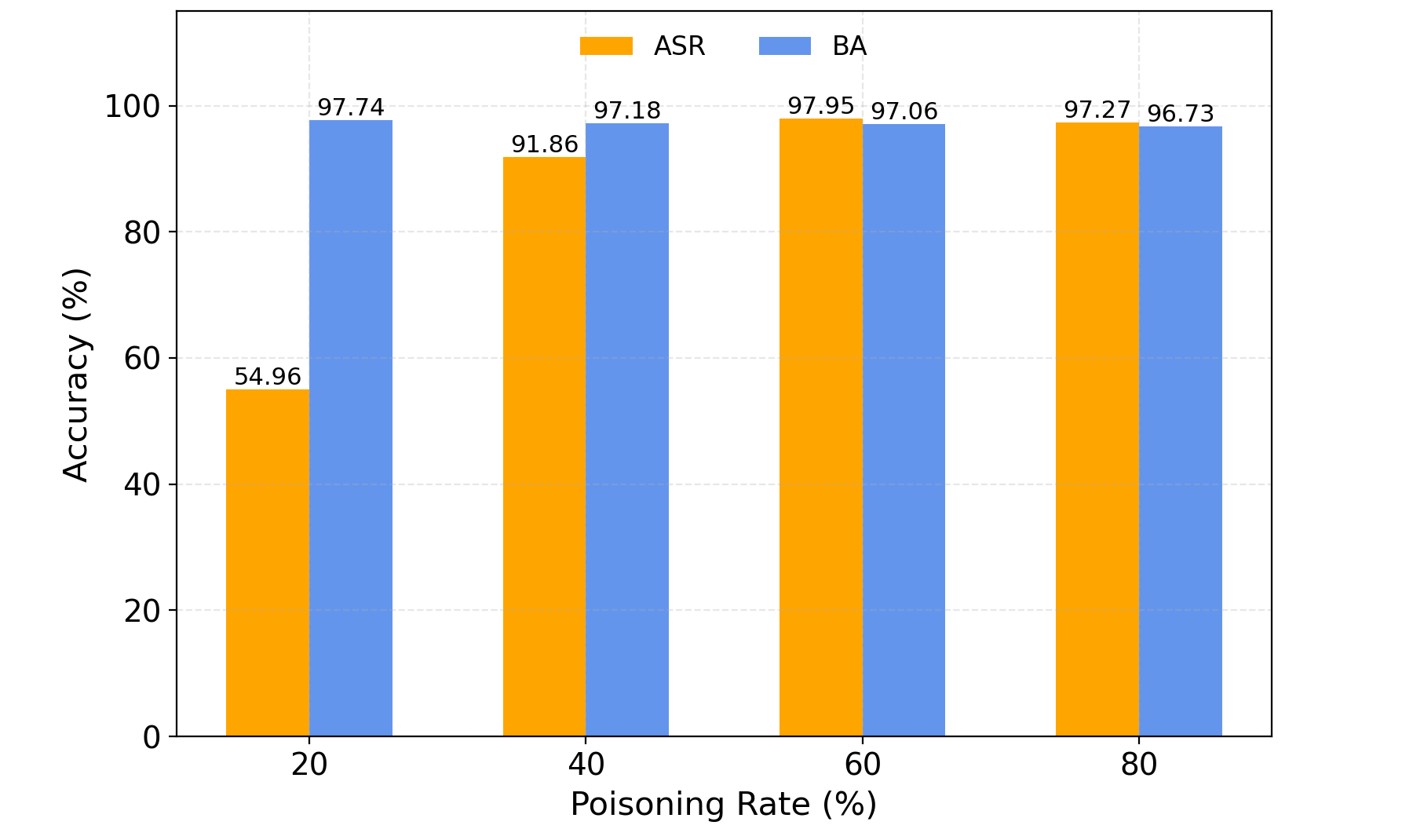}}\\
\subfigure[The ASR and BA w.r.t. different lengths of hash code.]{
 \includegraphics[width=0.975\columnwidth]{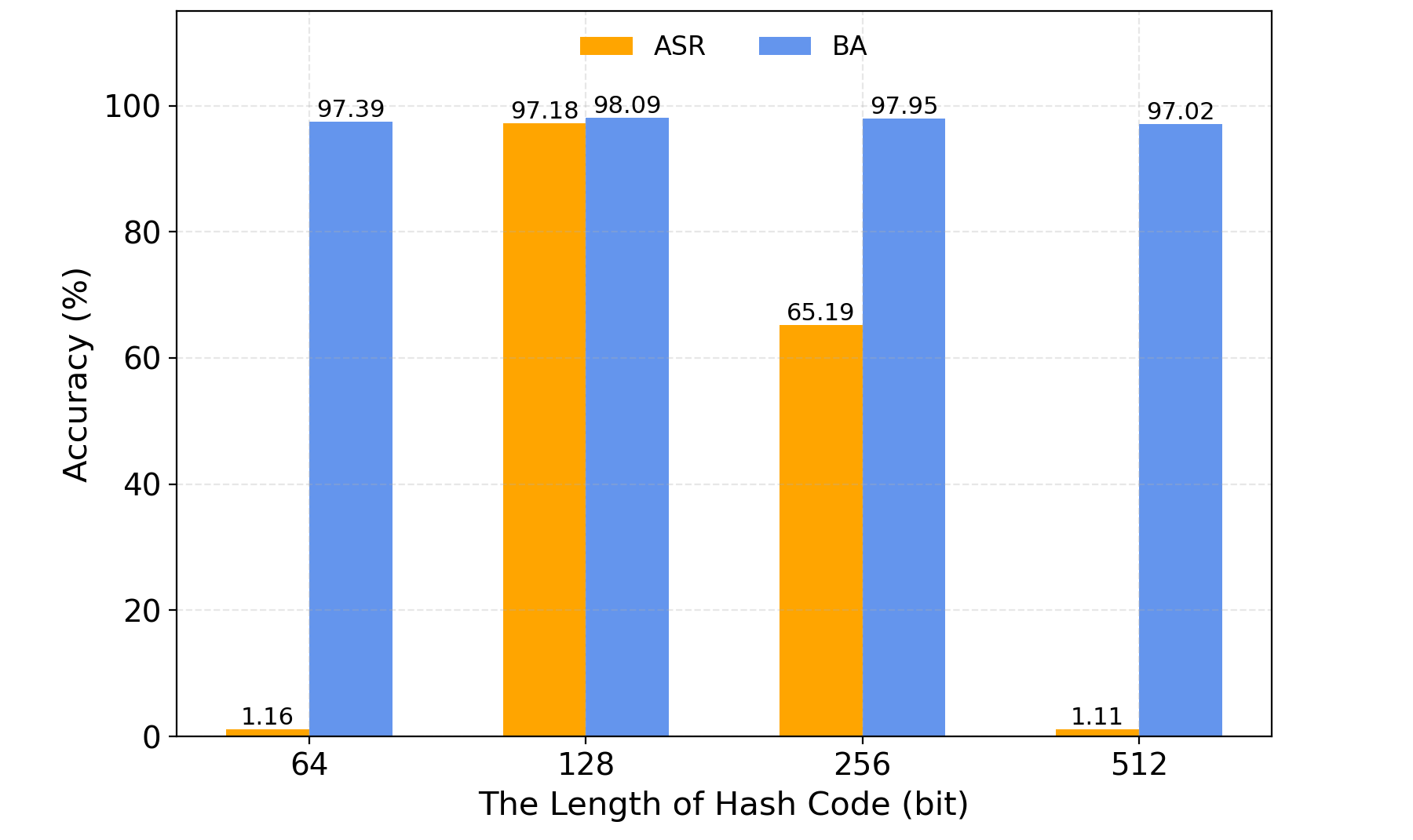}
}
\caption{Ablation experiments of the poisoning rate and the length of the hash code.}
\label{ablation}
\end{figure}

\paragraph{Poisoning Rate} Fig.~\ref{ablation} (a) shows the ASR and BA change over different poisoning rates, where both BA and ASR are affected by the poisoning rate.
Specifically, ASR increases as the poisoning rate rises, whereas BA maintains at or above 96.73\% .
The ASR is 91.86\% when the poisoning rate $\gamma = 38.8\%$, and reaches 96.94 \% when $\gamma = 75\%$. 
Note that an increase in the poisoning rate increases the likelihood that the backdoor would be discovered, and the adversary must strike a balance between attack effectiveness and stealthiness.

\paragraph{Length of Hash Code}

To generate backdoor triggers, we compress the high-dimensional identity feature of a reference image into a low-dimensional hash code. 
The length of the hash code correlates directly with the quality of the image generated by the image steganography network, as well as the identity information contained in the hash code, which consequently has an indirect impact on attack performance. 
Fig.~\ref{ablation}(b) shows the changes of BA and ASR over different hash code lengths at a poisoning rate $\gamma= 38.8\%$. 
We observe that our method achieves the highest ASR when the length of the hash code is 128 bits.
ASR drops significantly when the code length is larger than 128 because the embedded trigger damage the original information in the image. On the other hand, when the code length is small (64), ASR also decreases because the binary code too short does not contain enough identity information for the backdoor attack. 

\section{Conclusion}
Most of the current research on backdoor attacks focuses on image classification tasks, while the risk of backdoor attacks against person re-identification has rarely been studied. Existing backdoor attacks against image classification follow all-to-one/all scenarios and can not be directly applied to attack the open-set ReID model. 
As a result, we propose a novel backdoor attack on deep ReID models under a new all-to-unknown
scenario, which is able to dynamically generate new backdoor triggers containing unknown identities in the test set. Specifically,  an identity hashing network is proposed to first extract target identity information from a reference image, which is then injected into the benign images by image steganography. 
We show that the proposed attack method performs well in terms of effectiveness and stealthiness, and is robust to existing defense methods.
With some problems left open, we hope that this study will raise more attention on the backdoor attack risk against  person re-identification.

\bibliographystyle{IEEEtran}
\bibliography{ref}

\end{document}